\documentclass[10pt,journal,compsoc]{IEEEtran}

\ifCLASSOPTIONcompsoc
  \usepackage[nocompress]{cite}
\else
  \usepackage{cite}
\fi

\usepackage{mathptmx} 
\newcommand{\ignore}[1]{}
\usepackage{fancyhdr}
\usepackage[hyphens]{url}
\usepackage{mathptmx} 
\usepackage{microtype}
\usepackage[normalem]{ulem}
\usepackage{xspace}

\usepackage{xspace}
\usepackage{color,soul} 
\usepackage{ulem} 
\usepackage{graphicx}
\usepackage[table,xcdraw]{xcolor}
\usepackage{cite}
\usepackage[font={normalsize},labelfont={normalsize,bf}]{caption}

\usepackage[bookmarks=true,breaklinks=true,letterpaper=true,colorlinks,linkcolor=black,citecolor=blue,urlcolor=black]{hyperref}

\usepackage[utf8]{inputenc}
\usepackage{listings}
\usepackage{color}
\usepackage{enumitem}
\definecolor{codegreen}{rgb}{0,0.6,0}
\definecolor{codegray}{rgb}{0.5,0.5,0.5}
\definecolor{codepurple}{rgb}{0.58,0,0.82}
\definecolor{backcolour}{rgb}{0.95,0.95,0.92}

\newcommand{\precap}{}
\newcommand{\postcap}{}

\newcommand{\TensorOpt}{TVM\xspace}
\newcommand{\accel}{VTA\xspace}
\newcommand{\accelmeaning}{Versatile Tensor Accelerator\xspace}

\newcommand{\squishlist}{
   \begin{list}{$\bullet$}
    { \setlength{\itemsep}{1pt}      \setlength{\parsep}{3pt}
      \setlength{\topsep}{3pt}       \setlength{\partopsep}{0pt}
      \setlength{\leftmargin}{1em} \setlength{\labelwidth}{1em}
      \setlength{\labelsep}{0.5em} } }

\newcommand{\squishlisttwo}{
   \begin{list}{$\bullet$}
    { \setlength{\itemsep}{0pt}    \setlength{\parsep}{0pt}
      \setlength{\topsep}{0pt}     \setlength{\partopsep}{0pt}
      \setlength{\leftmargin}{2em} \setlength{\labelwidth}{1.5em}
      \setlength{\labelsep}{0.5em} } }

\newcommand{\squishend}{
    \end{list}  }

\begin{document}

\title{A Hardware-Software Blueprint for \\Flexible Deep Learning Specialization}

\author{
  \small{
  	Thierry Moreau$ ^1$, \ \ 
    Tianqi Chen$ ^1$, \ \ 
    Luis Vega$ ^1$, \ \ 
    Jared Roesch$ ^1$, \ \ 
    Eddie Yan$ ^1$, \ \ 
    Lianmin Zheng$ ^2$, \ \ 
    Josh Fromm$ ^1$, \ \
    Ziheng Jiang$ ^1$, \ \
 }\\
  \small{Luis Ceze$ ^1$, \ \ \ Carlos Guestrin$ ^1$, \ \ \ Arvind Krishnamurthy$ ^1$}\\
  \small{$ ^1$Paul G. Allen School of Computer Science \& Engineering, University of Washington}\\
  \small{$ ^2$Shanghai Jiao Tong University}
}

\maketitle
\pagestyle{plain}

\begin{abstract}
Specialized Deep Learning (DL) acceleration stacks, designed for a specific set of frameworks, model architectures, operators, and data types, offer the allure of high performance while sacrificing flexibility.
Changes in algorithms, models, operators, or numerical systems  threaten the viability of specialized hardware accelerators.

We propose VTA, a programmable deep learning architecture template designed to be extensible in the face of evolving workloads.
VTA achieves this flexibility via a parametrizable architecture, two-level ISA, and a JIT compiler.
The two-level ISA is based on (1) a task-ISA that explicitly orchestrates concurrent compute and memory tasks and (2) a microcode-ISA which implements a wide variety of operators with single-cycle tensor-tensor operations. Next, we propose a runtime system equipped with a JIT compiler for flexible code-generation and heterogeneous execution that enables effective use of the VTA architecture.

VTA is integrated and open-sourced into Apache TVM, a state-of-the-art deep learning compilation stack that provides flexibility for diverse models and divergent hardware backends.
We propose a flow that performs design space exploration to generate a customized hardware architecture and software operator library that can be leveraged by mainstream learning frameworks.
We demonstrate our approach by deploying optimized deep learning models used for object classification and style transfer on edge-class FPGAs.

\end{abstract}


\section{Introduction}

Hardware specialization is a powerful way to accelerate a known set of applications and workloads. Unfortunately, deep learning is
anything but a static field, and machine learning community rapidly changes the software they use to write models, the architecture of models themselves, the operators used by said models, and the data types they operate over.

The research community has primarily focused on two approaches for accelerator designs, fixed function accelerators 
and programmable accelerators (also known as Domain-Specialized Accelerators).
Current solutions offer compelling peak performance, but often fail to integrate into the evolving machine learning landscape.

Fixed-model accelerators are commonly spatially and statically laid out, offering attractive performance for certain workloads.
Unfortunately, the static nature of this approach rules out the reuse of hardware resources, limiting support for larger or newer models.

In contrast, programmable accelerators~\cite{Jouppi:TPU} offer far more flexibility by leveraging ISAs.
Due to the programmable nature of these accelerators, achieving peak performance requires a competent deep learning compiler that can map a large number of workloads onto a fixed set of hardware intrinsics.
Consequently, customizing behavior of these accelerators, even when open-sourced, is highly dependent on the availability of a transparent and modular software stack. 

A central challenge in prior work is linking innovations in specialization
to the rapidly changing software of machine learning. This challenge is not specific to computer architecture; it is present at all levels of the stack.
An end-to-end approach requires integration between frameworks, systems, compilers, and architecture in order to execute state of the art machine learning using hardware acceleration. Peak FLOPs only provide value if a programmer can access them.

We present \accel (\accelmeaning{}): an explicitly programmed architecture paired with a capable JIT compiler and runtime that can evolve in tandem with deep learning models without sacrificing the advantages of specialization.
\accel makes the following contributions:

\begin{itemize}
    \item A programmable accelerator design that exposes a two-level programming interface: a high-level task ISA to allow explicit task scheduling by the compiler stack, and a low-level microcode ISA to provide software-defined operational flexibility.
    In addition the \accel architecture is fully parameterizable: the hardware intrinsics, memories, and data types can be customized to adapt to the hardware backend requirements.
    \item An extensible runtime system for heterogeneous execution that performs JIT compilation of microcoded kernels to provide operational flexibility. The \accel runtime has allowed us, for instance, to extend the functionality of \accel's original computer vision-centric design to support operators found in style transfer applications without requiring any modifications to the hardware. 
    \item A schedule auto-tuning platform that can optimize data access and data reuse in order to rapidly adapt to changes in the underlying hardware and changes in workload diversity.
\end{itemize}

\begin{figure*}[!t]
\centering
\includegraphics[width=1\textwidth]{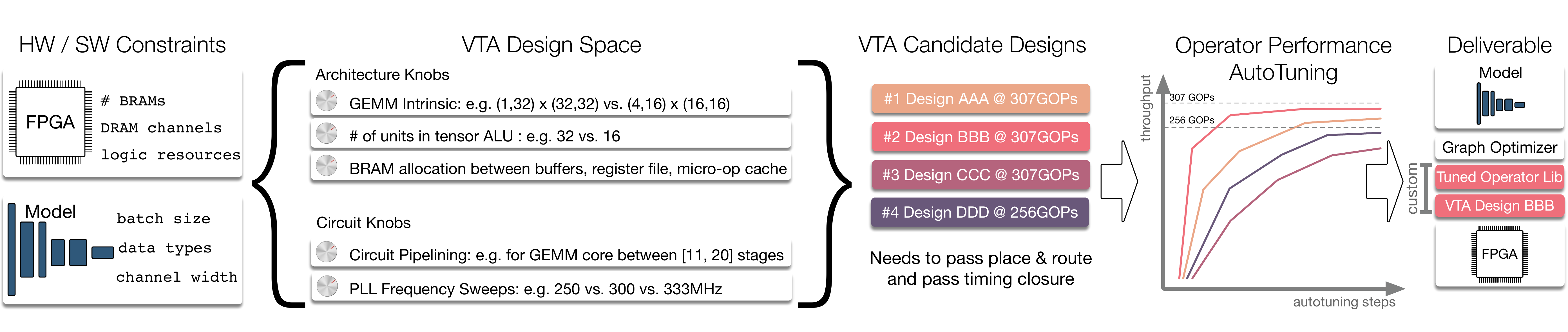}
\caption{\accel provides flexibility with respect to hardware targets and deep learning models. This flow diagram shows the steps in adapting a given model to a hardware backend by exploring \accel hardware configurations, and performing operator autotuning on the top hardware candidates. This process generates the pieces necessary to deploy \accel in any deep learning framework.}
\label{fig:evaluation_flow}
\end{figure*}

We demonstrate \accel's flexibility by adapting different workloads for two edge FPGAs. \autoref{fig:evaluation_flow} presents how to map a workload to FPGAs using the \accel architecture and runtime.
This process explores \accel hardware variants, and performs software autotuning for each candidate design. The resulting design and customized software binaries can be easily integrated into a deep learning framework.
Finally, we evaluate the full system, demonstrating \accel's ability to outperform edge GPUs with edge FPGAs on inference workloads.

\section{\accel Hardware-Software Stack Overview}
\label{sec:stack}

Running an end-to-end workload on \accel requires a complete software stack that can map high-level models down to the programming interface exposed by \accel.
We outline the layers of the \accel system stack below, which we built into the Apache \TensorOpt deep learning compiler stack.

\begin{figure}[t]
\centering
\includegraphics[width=1\columnwidth]{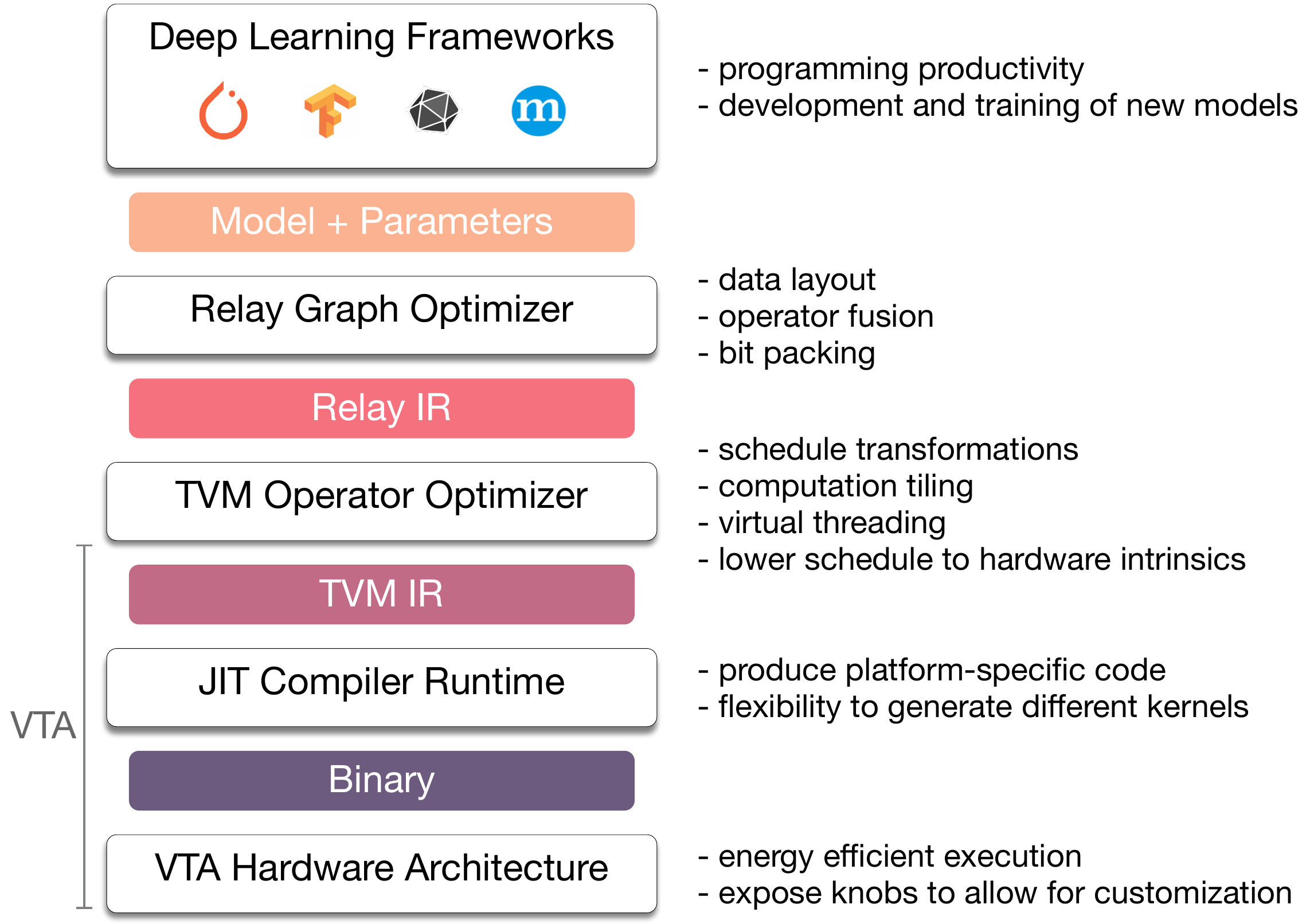}
\precap
\caption{Overview of the software stack built for \accel. We leverage the Apache \TensorOpt compiler stack to target \accel.}
\postcap
\label{fig:stack}
\end{figure}

\begin{description}
\item[Framework.]
Frameworks allow programmers to easily express models in a declarative fashion and perform training at scale on standard datasets.
Frameworks like TensorFlow, PyTorch, MxNet have gained widespread adoption, allowing the community to easily share, and deploy models.
TVM's ability to ingest models from these popular frameworks, 
enables generic compilation from frameworks to VTA.

\item[Relay Graph Optimizer.]
Relay~\cite{roesch2018relay} is TVM's high level program representation.
Relay generalizes the computation graphs used by prior frameworks and deep learning compilers into a full programming language.
The Relay optimization pipeline performs generic optimizations such as operator fusion and partial evaluation.
Relay's design is focused on extensibility, a property we use to 
extend Relay with a set of optimizations specific to \accel.
When targeting \accel we quantize inputs to match \accel's low precision data types, transform data layout, maximize data reuse, and transform input and weight data layouts to utilize \accel's tensor intrinsics.

\item[TVM Operator Optimizer.] 
TVM~\cite{Chen:TVM} automates the tedious process of scheduling workloads onto \accel accelerator variants.
Scheduling is important for multiple reasons.
First, it tiles the computation to maximize data reuse.
Second, it inserts thread parallelism that \accel's runtime can translate into task-level pipeline parallelism.
Third, it partitions operators into sub-computations which can be mapped to high-level hardware intrinsics such as bulk DMA load or GEMM.
TVM incorporates AutoTVM~\cite{chen2018learning}, an automated schedule optimizer.
We rely upon AutoTVM to guide our hardware candidate exploration search for the best \accel candidates given a workload.

\item[JIT Compiler and Runtime.]
The runtime performs JIT compilation of the accelerator binaries and manages heterogeneous execution between the CPU and \accel.
The JIT compiler abstracts binary compatibility by introducing one level of indirection.
We describe the runtime in more details in Section~\ref{sec:runtime}.

\item[Hardware architecture.]
\accel is a parameterizable accelerator that accelerates the bulk of the deep learning compute graph.
\accel is explicitly programmed by the compiler stack using a two-level programming interface.
The architecture is parameterized by the size of the GEMM core, the SRAM shapes, and data type widths.
A parameterized hardware architecture makes it possible to retarget the same design to devices with different hardware resources.
We describe \accel in more details in Section~\ref{sec:arch_overview}.
\end{description}

\begin{figure}[t]
\centering
\includegraphics[width=1\columnwidth]{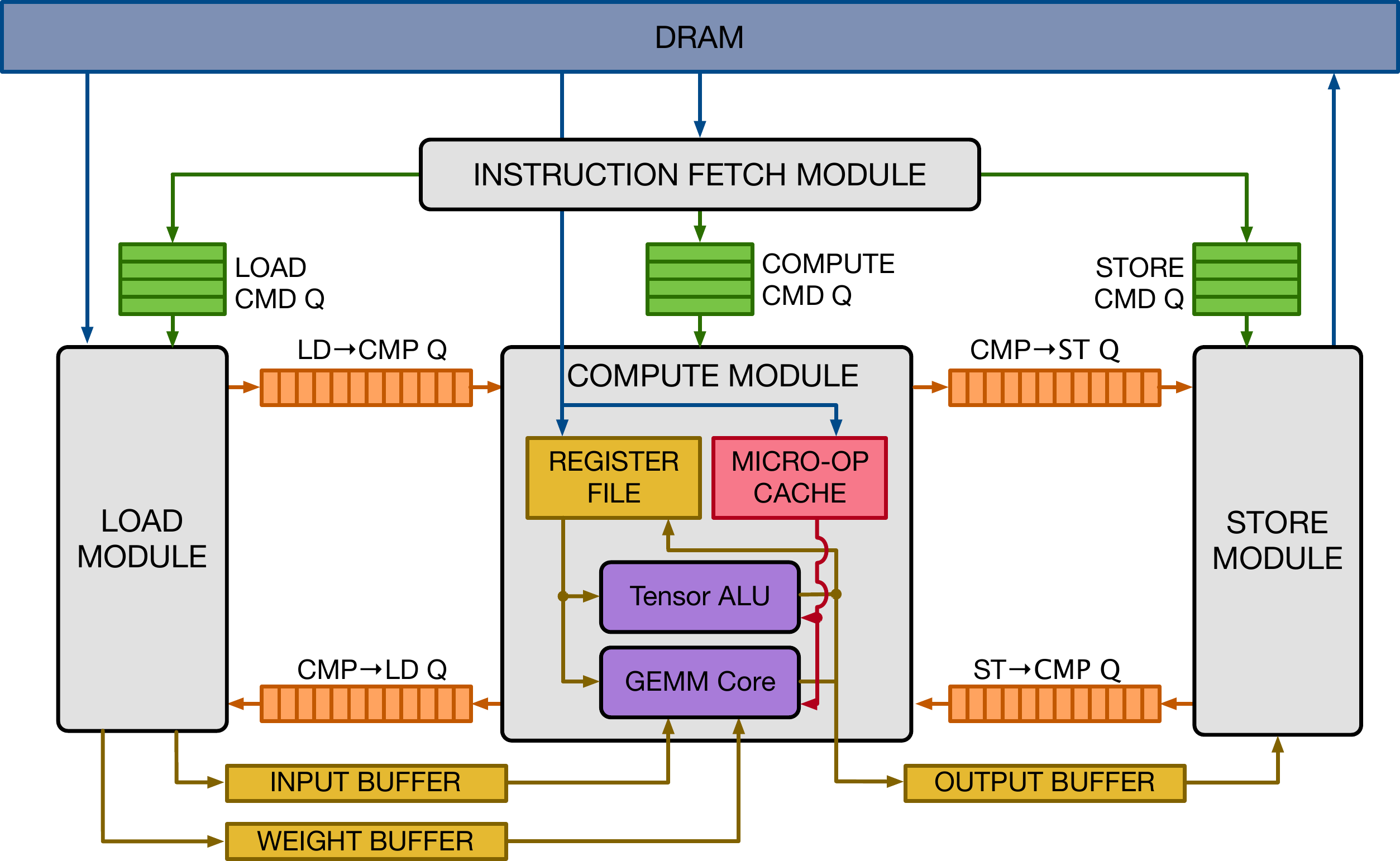}
\precap
\caption{The \accel hardware organization. \accel is composed of modules that communicate via queues and SRAMs. This defines a task pipeline, which helps maximize compute resource utilization.}
\postcap
\label{fig:hw_overview}
\end{figure}


\begin{figure*}[t]
\centering
\includegraphics[width=1\textwidth]{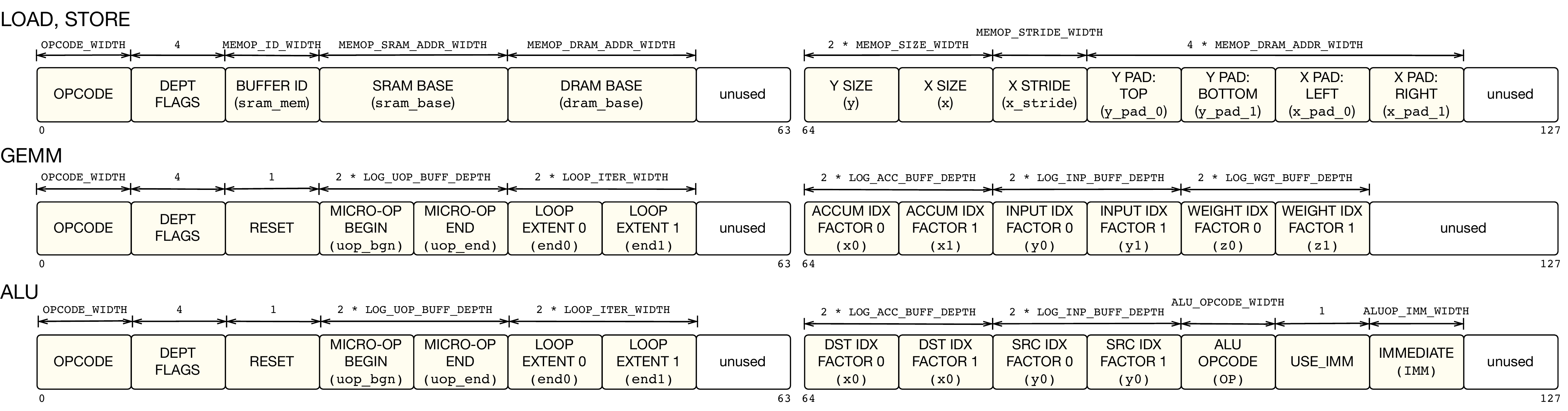}
\precap
\caption{The \accel high-level instruction fields. \texttt{LOAD} and \texttt{STORE} instructions perform 2D strided DMA reads/writes between DRAM and SRAM. \texttt{GEMM} instructions are used to matrix multiplication, 2D convolutions, etc. while \texttt{ALU} instructions can perform a wide range of activation, normalization, and pooling tasks.}
\postcap
\label{fig:hw_instructions}
\end{figure*}

\section{\accel Architecture and JIT Runtime}
\label{sec:hw}

A successful implementation of a flexible deep learning accelerator requires co-design of the hardware with the software stack.
We describe at a high level two components that were co-designed to achieve this goal: the \accel hardware architecture, and the \accel JIT compiler and runtime.

\subsection{Hardware Architecture}
\label{sec:arch_overview}
\autoref{fig:hw_overview} gives a high-level overview of the \accel hardware organization.
\accel is composed of four modules: \texttt{fetch}, \texttt{load}, \texttt{compute}, and \texttt{store}.
Together, these modules define a \emph{task pipeline}, which enables high compute resource utilization on compute-bound workloads, and high memory bandwidth utilization on memory-bound workloads.
These modules communicate over command queues and on-chip shared memories (SRAMs) that act as uni-directional data channels.
Accesses to these memories are synchronized via dependency queues to prevent data hazards such as Read After Write (RAW) and Write After Read (WAR). Finally, the 3-stage architecture  (\texttt{load-compute-store}) can be used to build task pipelines of arbitrary depth as long as dependencies are properly managed.

\noindent{\textbf{Parameterizability.}} The \accel architecture is fully parameterizable:
the shape of the GEMM tensor intrinsic can be modified to influence the utilization of hardware resources.
Modifying the shape of the input, weight, and accumulator tensors that feed the GEMM unit directly affects how many multipliers to instantiate and how wide SRAMs ports need to be.
In addition, each data type can customized to a different integer precision: weight and input types can be 8-bits or fewer, while the accumulation type can be 32-bits or fewer.
Control of integer precision lets us scale arithmetic density on chip when resources are constrained.

\noindent{\textbf{Exposing Task-Level Pipeline Parallelism.}}
Task-Level Pipeline Parallelism (TLPP) is an important feature in the \accel architecture, because it enables simultaneous use of compute and memory resources to maximize their utilization.
TLPP is based on the paradigm of access-execute decoupling~\cite{Smith:DAE}.
To extract TLPP, we partition tasks into two mutually-exclusive execution contexts, so that concurrent load, compute, and store operations do not interfere with one another.
This partitioning is easily achieved in TVM using virtual threads~\cite{Chen:TVM}.
To guarantee timely and correct execution for decoupled access-execute instruction streams, we encode dependency information into instructions.
This effectively results in memory latency hiding on compute-bound workloads (e.g. 2d convolutions).

\noindent{\textbf{Task-Level ISA.}}
\accel supports a high-level task ISA that encodes multi-cycle compute and memory operations, including \texttt{LOAD}, \texttt{GEMM}, \texttt{ALU}, and \texttt{STORE} instructions described in Figure~\ref{fig:hw_instructions}.
\texttt{LOAD} and \texttt{STORE} instructions describe how data from DRAM is loaded and stored into on-chip SRAMs.
Strided memory access is supported to load tensor tiles without  modifying memory layout.
\texttt{GEMM} and \texttt{ALU} instructions invoke micro-coded kernels, based on micro-op instructions, which describe the data-access patterns that define a given deep learning operator.

We illustrate a simple execution pipeline in VTA below:
\squishlist
\item The \texttt{fetch} module loads task instructions from DRAM and dispatches them, according to the instruction type, to the corresponding command queues connected to \texttt{load}, \texttt{compute}, and \texttt{store} modules.
\item The \texttt{load} module loads input, weight, and bias tensor tiles from DRAM into on-chip memories.
\item The \texttt{compute} module loads a micro-coded kernel from DRAM into on-chip memory. Micro-coded kernels are based on micro-ops that describe data access patterns for inputs, weights, and biases.
\item The \texttt{compute} module executes the micro-coded kernel to perform either a dense linear algebra computation via the GEMM core or a pairwise arithmetic operations via the Tensor ALU.
\item The \texttt{store} module reads results processed by the \texttt{compute} module and writes them to DRAM.
\squishend

\begin{figure*}[t]
\centering
\includegraphics[width=0.80\textwidth]{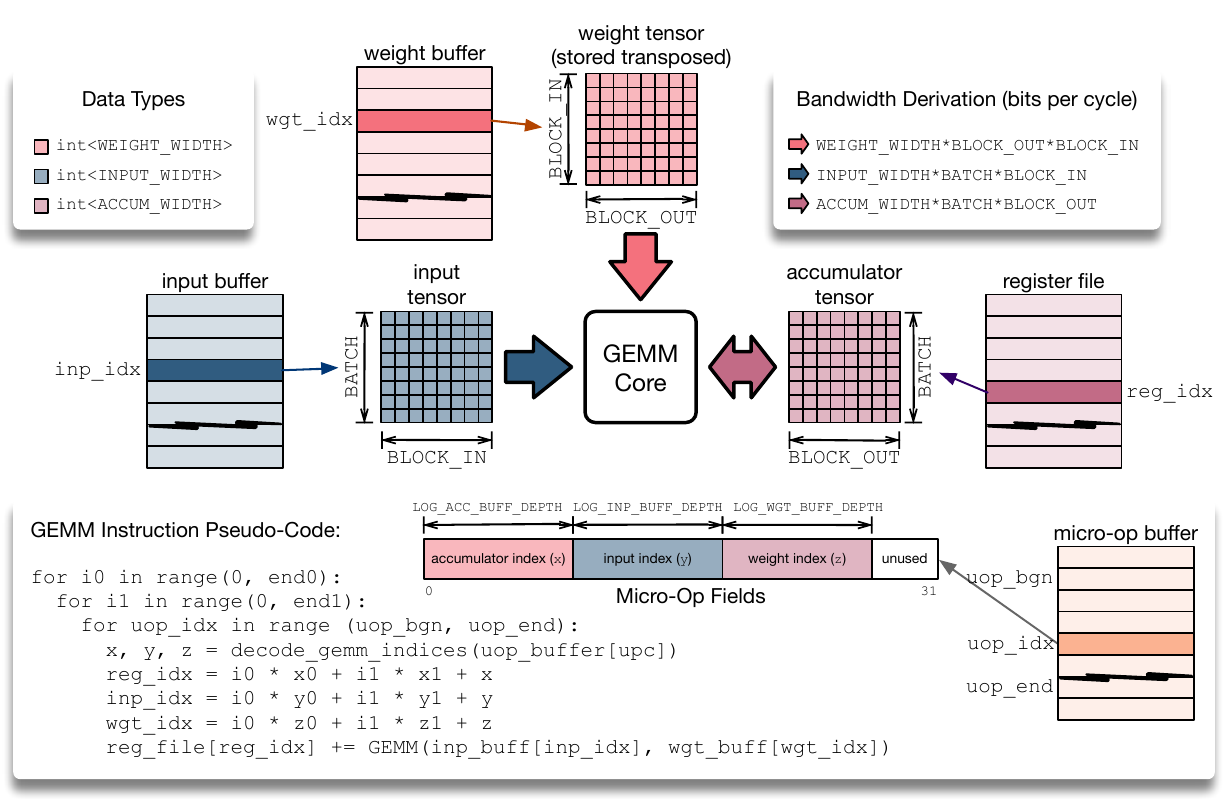}
\precap
\caption{The GEMM core can perform one dense matrix multiplication over an input tensor and a weight tensor, and add the result into a register file tensor. The data addressing pattern is specified by a micro-coded sequence: this allows us to map different deep learning operators onto a single fixed-sized matrix-matrix multiplication intrinsic.}
\postcap
\label{fig:gemm_core}
\end{figure*}

\noindent{\textbf{Compute Module.}}
Two functional units perform operations on the register file: the tensor ALU and the GEMM core.
The tensor ALU performs element-wise tensor operations such as addition, activation, normalization, and pooling tasks.
The GEMM core performs high-arithmetic intensity matrix multiplication over input and weight tensors to implement common deep learning operators such as 2D convolutions, or fully connected layers.

The GEMM core performs matrix multiply operations at a pipelined rate of one input-weight matrix multiplication per cycle.
Its logic is implemented as parallel vector dot-product using reduction trees, but can be substituted with other implementations such as systolic arrays.
The GEMM core defines a low-level tensor \emph{hardware intrinsic} which is exposed to the \TensorOpt compiler stack.
\TensorOpt uses \emph{tensorization}~\cite{Chen:TVM}: an automated approach to mapping deep learning operators such as 2d convolution down to fixed tensor hardware intrinsics.

\noindent{\textbf{Microcode ISA.}}
The \texttt{compute} core reads instructions from the micro-op cache, which describe how computation is performed over data.
\autoref{fig:gemm_core} details how the GEMM core performs computation over data stored in the input, weight, and accumulator memories.
These micro-ops provide no control flow.
Therefore, instructions need to be unrolled to express repeatable data access stencils.
There are two types of compute micro-ops: ALU and GEMM operations.
To minimize the footprint of micro-op kernels while avoiding the need for control-flow instructions, the compute core executes micro-op sequences inside a two-level nested loop that computes the location of each tensor register via an affine function.
This compression approach helps reduce the micro-ops footprint when sent to the accelerator.

\subsection{JIT Runtime System}
\label{sec:runtime}
\accel's JIT runtime enables cooperative execution of deep learning workloads between a CPU host and the accelerator.
The JIT runtime design follows five objectives:
(1) enable heterogeneous execution, (2) lower compiler design complexity, (3) overcome physical limitations, (4) reduce binary bloat, (5) future proofing.

\noindent{\textbf{Heterogeneous execution.}}
One challenge present in fixed function accelerators is model evolution, because most of these accelerators are built for fixed models. Heterogeneous execution overcomes this limitation by properly scheduling operators into targets (e.g., CPUs or \accel), depending on their affinity for different types of operators. For instance, it is well known that the first convolutional layer in most CNNs contains operators with low arithmetic intensity that perform well on CPUs. Another motivation behind heterogeneous execution is providing a \emph{fallback} mechanism for supporting emerging operators that are not yet supported by \accel.

\noindent{\textbf{Compiler Design.}}
By adding a level of indirection, code JIT-ting eliminates the need to write compiler code-generation backends which can be tedious to maintain for different programmable accelerators.
The JIT compiler exposes a high-level API to TVM to lower schedules onto, abstracting away \accel variant-specific architectural details.
This lets us extend the TVM compiler support we built for \accel to cover future variants of different shapes and sizes.

\noindent{\textbf{Physical Limitations.}} The JIT runtime generates and manages micro-kernels on the fly.
It controls when to load kernels from DRAM into the accelerator limited micro-op cache.
This eliminates micro-op memory physical limitations and lets us support large models, even if all micro-kernels for all layers do not fit in SRAM all at once.
It also lets us trade area used by the micro-op cache for other resources such as data storage, or compute units.

\noindent{\textbf{Binary bloat.}}
Delaying micro-kernel generation to the JIT compilation stage minimizes binary bloat.
Since \accel's architecture has limited support for control flow, micro-kernels have to be unrolled which can produce fairly large binaries.
In addition, micro-kernel code JIT-ting expresses binaries for heterogeneous execution in a single-ISA: instead of shipping a hybrid binary, we just ship one CPU binary that performs accelerator binary JIT-ting.

\noindent{\textbf{Future proofing.}}
Advancements in deep learning have outlined the prevalence of dynamic neural network workloads that incorporate control flow.
Additionally, advances in systems show trends towards heterogeneous multi-accelerator systems and scale-out acceleration.
Having a runtime that handles dynamic decisions across heterogeneous platforms will keep the design of hardware accelerators like \accel simple, and constrain support for future models to being a mostly software-defined problem.


\section{VTA Hierarchical Optimization}

\begin{figure}[t!]
\centering
\includegraphics[width=1\columnwidth]{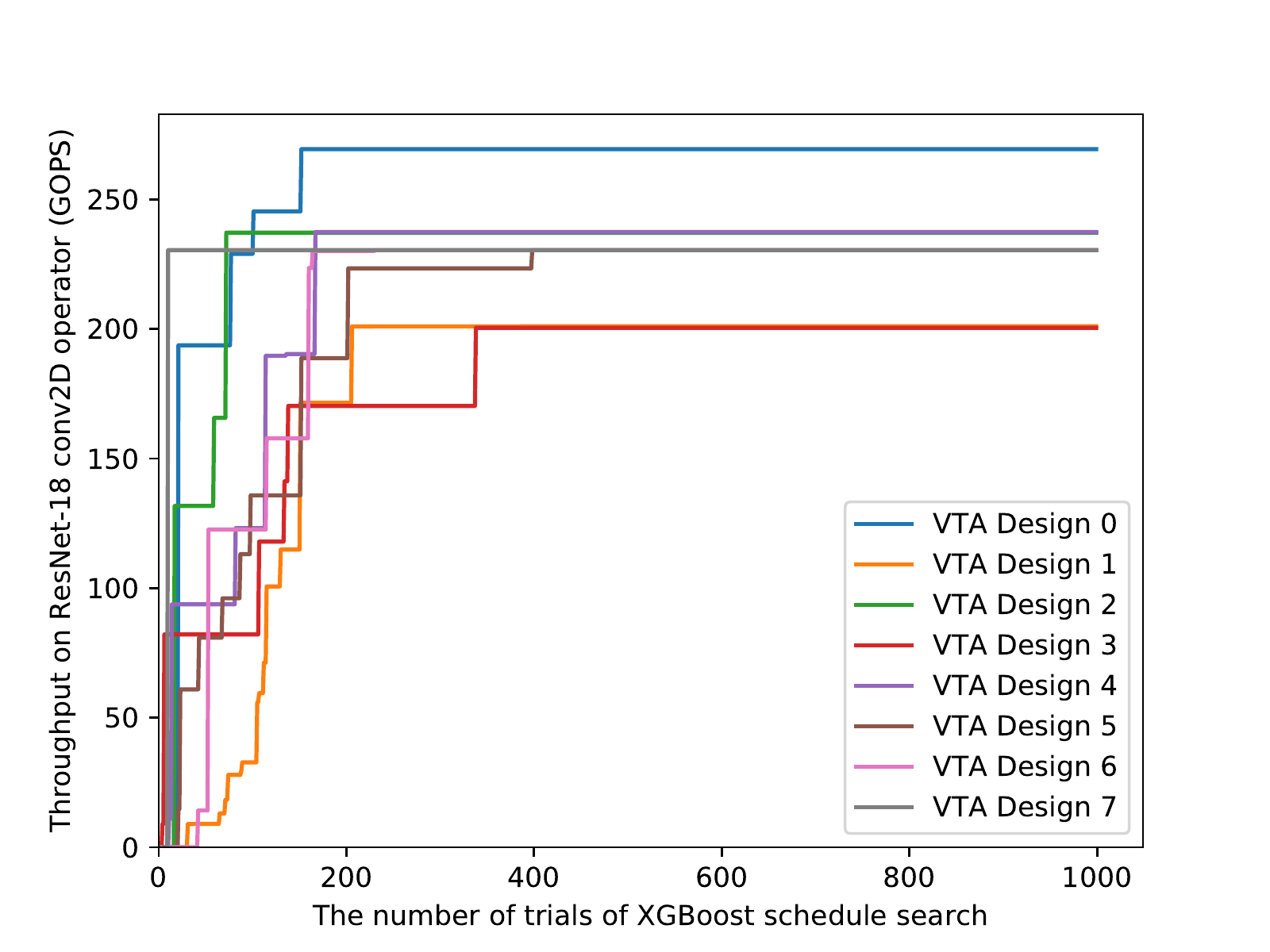}
\caption{Schedule exploration with XGBoost for a single ResNet-18 layer on Ultra-96. Eight \accel design candidates with (2,16)x(16,16) and (8,8)x(8,8) GEMM intrinsic at W8A8 are considered. Layer is conv2d: IC=256, OC=256, H=W=14, KW=KH=3, stride=(1,1),  padding=(0,0).}
\label{fig:autotuning}
\end{figure}

\subsection{Hardware Exploration for Varying FPGA Sizes}

One way to showcase \accel's architecture flexibility is to target different FPGA platforms.
FPGAs are becoming more accessible than ever, with sub-\$100 development boards, and FPGA cloud computing instances becoming ubiquitous.

The \accel design offers multiple architectural customization parameters that are listed in \autoref{fig:evaluation_flow}.
Architectural knobs include GEMM hardware intrinsic shape, data types, number of parallel arithmetic units in the tensor ALU, ALU operations, BRAM distribution between on-chip memories.
Circuit knobs include degree of hardware pipelining to close timing at higher frequencies, and PLL frequency.
These customization knobs define a hardware design space with 100s to 1000s of individual designs.
This design space can be exhaustively explored to find the best candidate for a particular workload. We perform this exploration in a sequence of stratified steps. First we use a simple FPGA resource model to prune infeasible VTA parameterizations. After pruning, each candidate hardware design is compiled, placed, and routed. 
We pick the best feasible design for each $\{{fpga} \times {dtype} \times {batch}\}$ combination, but typically our exploration returns a handful of promising candidates -- the rest of the designs either yield low peak performance or fail placement, routing, or timing closure. For this final set of designs, we generate optimized software, using operator autotuning~\cite{chen2018learning}, and use this software to obtain the workload's performance profile.

An analytical model of \emph{peak performance} is used to initially filter hardware designs based on theoretical throughput and frequency assuming compute resources are 100\% utilized.
However, assuming 100\% utilization of compute resources by a particular operator is often inaccurate.
For example, depending on the workload mix, operators like \texttt{conv2d} with large window sizes may exhibit high arithmetic intensity (measured in Op/Byte). Operations with high arithmetic intensity translate to high utilization, and therefore are close to peak performance.
Operators which exhibit low arithmetic intensity, like \texttt{conv2d}, with a window size of 1, are memory bandwidth constrained. In these situations we are able to use task-level pipeline parallelism to mitigate performance loss.

\begin{figure}[t!]
\includegraphics[width=1\columnwidth]{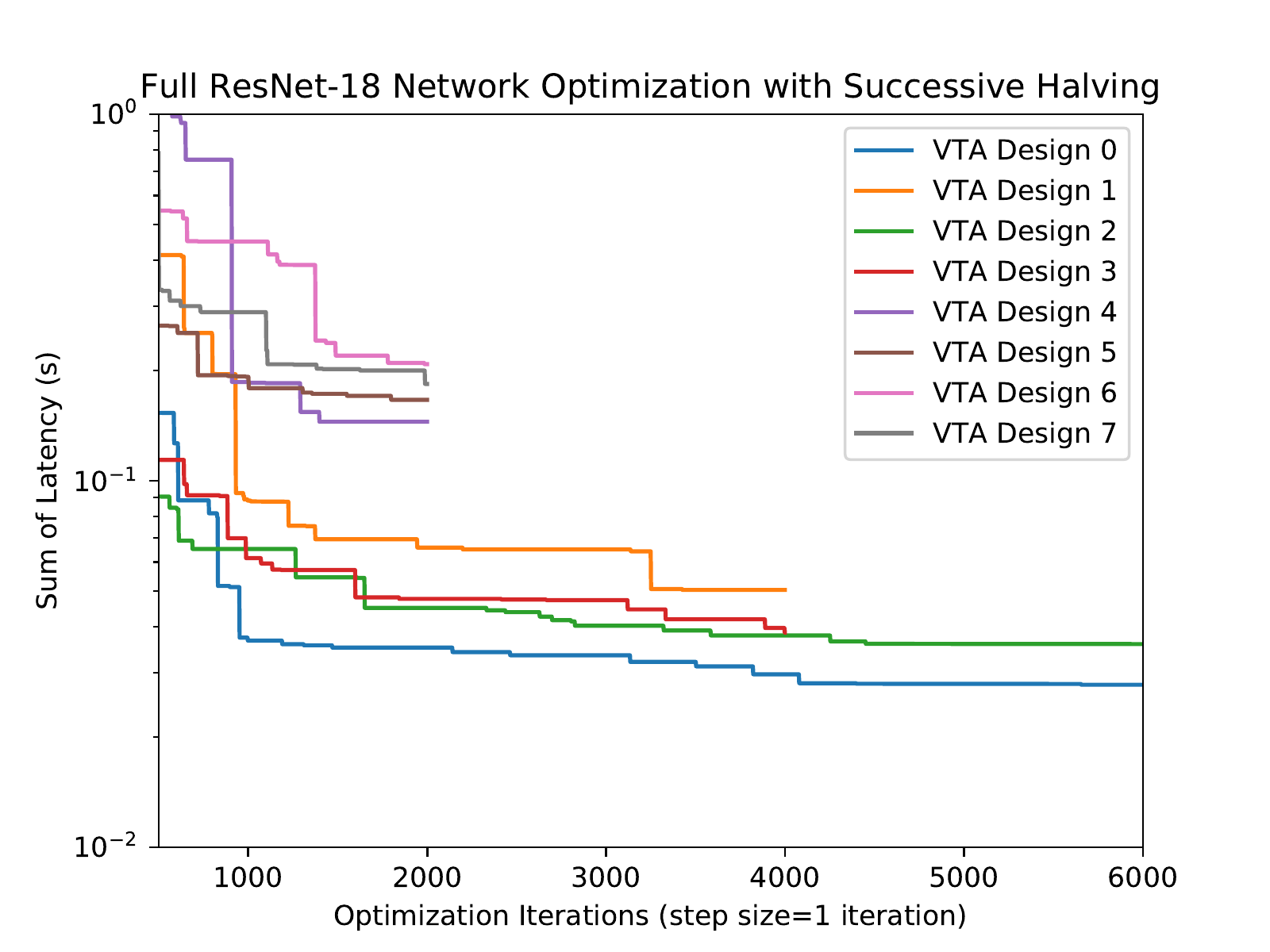}
\caption{Example of hardware design exploration and schedule autotuning on a complete ResNet-18 inference workload run on Ultra96 FPGA. The exploration begins with promising \accel hardware variants and converges to the optimal hardware design while using a fraction of the optimization time required to exhaustively evaluate each hardware design. }
\label{fig:vtahierarchy}
\end{figure}

\subsection{Schedule Exploration for Operator Autotuning}
\label{sec:eval_sched}

Schedule autotuning is the process by which an automated search algorithm attempts to optimize a given program or workload towards peak hardware performance. We perform autotuning by applying different memory tiling, loop transformations (e.g. splitting, reordering, unrolling), vectorization/tensorization, and parallelization strategies~\cite{chen2018learning}.
We then use the TVM compiler to express schedule templates for each operator (e.g. \texttt{conv2d}, \texttt{conv2d\_transpose}, \texttt{group\_conv2d}, \texttt{fc}) we support in hardware. We use TVM's automated scheduling library to obtain schedules that maximize performance for a given combination of operator, tensor shape, and hardware parameterization.

\autoref{fig:autotuning} shows the autotuning search process when optimizing different \accel hardware candidates for a single ResNet layer.
We used the XGBoost~\cite{Chen:Xgboost} search algorithm to find the best schedules for each hardware variant in a limited number of trials. Each workload's layers are then tuned for each hardware candidate. Aggregate inference time is used to 
select the \accel hardware variant that is best for a given model.

It takes several hours to exhaustively tune a network on a single hardware variant.
Given the large number of \accel hardware designs to test, and model architectures to support, autotuning search quickly becomes intractable without careful design.
Minimizing full-network autotuning time across multiple hardware candidates introduces a \emph{hierarchical prioritization problem}. We approach this challenge by applying a hyperparameter optimization technique, based on \texttt{SuccessiveHalving}~\cite{jamieson2016non}.
Instead of choosing among hyperparameters that define a network architecture, we apply this technique to choose among \accel design candidates. We simultaneously inspect how the relative performance of each hardware design evolves for a given workload, over each iteration of the optimization algorithm. Throughout optimization we use a round-robin policy to update latency estimates across all operators for each hardware design.

\subsection{Full Network Optimization Case Study}

We show in \autoref{fig:vtahierarchy} an example of hierarchical optimization for the ResNet-18 workload, based on the hardware exploration and schedule exploration techniques described before. We perform these optimizations over a set \accel candidates generated using W8A8 (8-bit weights, 8-bit activations) data representations. We select eight promising hardware candidates, and apply \texttt{SuccessiveHalving} to prune designs that do not appear promising.
Similar to hyperparameter optimization for neural network training, this is a difficult task, as the relative performance differences between hardware designs may be small early on. After a moderate number of iterations, \texttt{SuccessiveHalving} is able to converge to the best candidate hardware design.

This case study showcases \accel's ability to quickly navigate a non-trivial space of accelerator configurations for a given workload.
As accelerator configurations change, so does the software that programs it.
This joint-optimization problem can only be solved with a flexible stack.

\section{Evaluation}
\label{sec:eval}

As the landscape of deep learning continues to evolve, it is important to support emerging models.
We evaluate \accel's ability to support two novel model architectures beyond standard deep convolution nets.
First, we evaluate MobileNet, a recent model architecture that uses grouped convolution to reduce the total computation overhead of the network.
We evaluate a variant of MobileNet we call MobileNetG that groups channels by the vector factor of the \accel's GEMM core.
Second, we implement a Generative Adversarial Network (DCGAN) model that is used for image-to-image translation and generation.

Both models require non-trivial extensions to support new operators.
MobileNetG requires support for grouped convolutions that exhibit block sparse patterns on channel groups.
DCGAN requires support for 2D convolution transpose which has a spatial sparsity pattern.
Accelerators must support these access patterns to avoid unnecessary computations and achieve maximum performance. The runtime can readily make use of schedules to generate micro-kernels that support these access patterns without changing the hardware.

We integrated \accel into Apache TVM and evaluated a variety of deep learning models on a set of edge FPGA devices with different resource budgets. We imported all models from MxNet~\cite{MXNet-whitepaper} a deep learning framework used by Amazon.
It is worth noting Relay's model importers provide access to a wide variety of other front-ends, and VTA is not limited to MxNet.

\begin{figure}
\centering
\includegraphics[width=1\columnwidth]{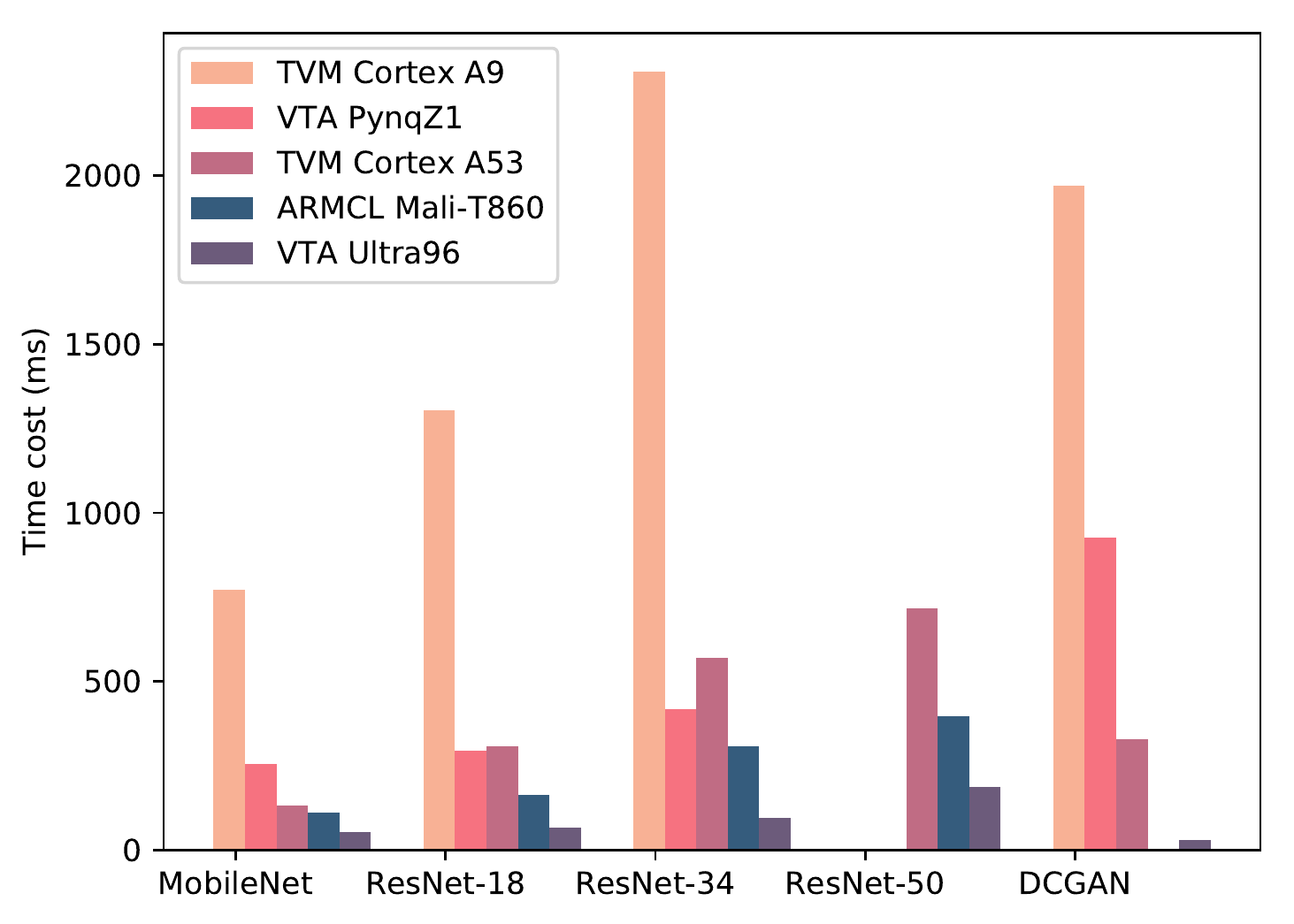}
\caption{End to end performance evaluation over multiple CPU, GPU and FPGA-equipped edge systems. For comparable systems, \accel provides a significant performance edge over conventional CPU and GPU-based inference.}
\label{fig:cpu_fpga_e2e}
\end{figure}

\autoref{fig:cpu_fpga_e2e} shows a performance comparison across these models, comparing \accel-accelerated execution against a highly optimized ARM CPU and GPU platforms that rely on industry-strength deep learning libraries: ARM ComputeLib (ARM CL) and TVM.
The ARM Cortex-A9, ARM Cortex-A53, and Mali-T860 GPU are taken from the Pynq-Z1 (\$65), Ultra-96 (\$250), and the Firefly-RK3399 (\$200) boards.
For the \accel hardware designs, we use an automated 8-bit integer scaling and translation pass from 32-bit floating-point (FP32) with negligible accuracy degradation.
For our CPU baselines, we use the TVM autotuner to obtain FP32 CPU kernels that take advantage of NEON vectorization, multi-threading and state of the art scheduling tricks (spatial tiling, Winograd transform etc.).
For our GPU baseline, we use the ARM CL v18.03 and exploit 16-bit floating-point (FP16) library support.
ARM CL is missing support \texttt{conv2d} transpose for DCGANs, demonstrating \accel's ability to stay ahead of the curve for unconventional workloads.

\autoref{fig:cpu_fpga_e2e} shows end-to-end results that can be discussed in two groups of comparable devices in terms of cost: (1) \accel on the Pynq vs. Cortex-A9 (sub-\$100), and (2) \accel on Ultra96 vs. Cortex-A53 and Mali-T860 GPU (\$200-\$250).
First, \accel on the Pynq-Z1 outperforms the Cortex-A9 CPU by
$3.0\times$, $4.4\times$, $5.3\times$ and $2.1\times$ on MobileNet, ResNet-18, ResNet-34 and DCGAN.
Second, \accel on the Ultra-96 outperforms the Cortex-A53 by $2.5\times$, $4.7\times$, $6.0\times$, $3.8\times$ and $11.5\times$ on MobileNet, ResNet-18, ResNet-34, ResNet-50 and DCGAN.
In addition, \accel on the Ultra-96 outperforms the mobile-class Mali-T860 GPU by $2.1\times$, $2.5\times$, $3.2\times$ and $2.1\times$ on MobileNet, ResNet-18, ResNet-34 and ResNet-50. 

Overall, \accel demonstrates that the flexibility of the architecture can offer high performance while forming a evolutionary path forward for accelerating diverse workloads on various devices.

\section{Conclusion}
In this paper, we present hardware-software blueprint for ``flexible specialization'':
the idea that efficiency gains from hardware specialization is not mutually exclusive with workload flexibility. 
We present VTA, a parametrizable deep learning architecture that is explicitly programmed via a two-level ISA.
We co-design the accelerator with a runtime system that JIT compiles micro-kernels to provide operational flexibility.
With this approach, we support less conventional operators such as convolution transpose, and grouped convolutions without needing to apply changes to the hardware.
We show in our evaluation that \accel can effectively target different FPGAs, multiple workloads, and leverage off the shelf deep learning compilers to quickly integrate optimized software with specialized hardware.
Finally, we demonstrate that a well integrated hardware and software stack lets us perform full stack optimization and exploration on FPGAs.

\bibliography{ieee-micro-2019}
\bibliographystyle{plain}

\end{document}